\documentclass[11pt,a4paper]{article}
\usepackage[hyperref]{emnlp2018}
\usepackage{latexsym}

\usepackage{url}
\usepackage{graphicx}
\usepackage{booktabs}
\usepackage{amsmath}
\usepackage{multirow}
\usepackage{tabulary}

\aclfinalcopy 

\newcommand{\tsm}{$TSM$}

\newcommand{\vect}[1]{\boldsymbol{#1}}
\newcommand{\comment}[1]{}
	
\title{Topic-Specific Sentiment Analysis Can Help Identify Political Ideology}

\author{Sumit Bhatia \\
  IBM Research AI \\
  New Delhi, India \\
  {\tt sumitbhatia@in.ibm.com} \\\And
  Deepak P \\
  Queen's University Belfast \\
  Belfast, UK \\
  {\tt deepaksp@acm.org} \\}

\date{}

\begin{document}
\maketitle

\begin{abstract}
Ideological leanings of an individual can often be gauged by the sentiment one expresses about different issues. We propose a simple framework that represents a political ideology as a distribution of sentiment polarities towards a set of topics. This representation can then be used to detect ideological leanings of documents (speeches, news articles, etc.) based on the sentiments expressed towards different topics. Experiments performed using a widely used dataset show the promise of our proposed approach that achieves comparable performance to other methods despite being much simpler and more interpretable.
\end{abstract}

\section{Introduction}
\label{sec:intro}

The ideological leanings of a person within the \textit{left-right} political spectrum are often reflected by how one feels about different topics and by means of preferences among various choices on particular issues. 
For example, a left-leaning person would prefer nationalization and state control of public services (such as healthcare) where privatization would be often preferred by people that lean towards the right. Likewise, a left-leaning person would often be supportive of immigration and will often talk about immigration in a positive manner citing examples of benefits of immigration on a country's economy. A right-leaning person, on the other hand, will often have a negative opinion about immigration.

Most of the existing works on political ideology detection from text have focused on utilizing bag-of-words and other syntactic features to capture variations in language use~\cite{ideological-proportion-speeches,ideologyPrediction-text,rnn-ideology}. We propose an alternative mechanism for political ideology detection based on sentiment analysis. We posit that adherents of a political ideology generally have similar sentiment toward specific topics (for example, right wing followers are often positive towards free markets, lower tax rates, etc.) and thus, \emph{a political ideology can be represented by a characteristic sentiment distribution over different topics} (Section~\ref{sec:approach}). This topic-specific sentiment representation of a political ideology can then be used for automatic ideology detection by comparing the topic-specific sentiments as expressed by the content in a document (news article, magazine article, collection of social media posts by a user, utterances in a conversation, etc.).

In order to validate our hypothesis, we consider exploiting the sentiment information towards topics from archives of political debates to build a model for identifying political orientation of speakers as one of right or left leaning, which corresponds to republicans and democrats respectively, within the context of US politics. This is inspired by our observation that the political leanings of debators are often expressed in debates by way of speakers' sentiments towards particular topics. Parliamentary or Senate debates often bring the ideological differences to the centre stage, though somewhat indirectly. Heated debates in such forums tend to focus on the choices proposed by the executive that are in sharp conflict with the preference structure of the opposition members. Due to this inherent tendency of parliamentary debates to focus on topics of disagreement, the sentiments exposited in debates hold valuable cues to identify the political orientation of the participants. 

We develop a simple classification model that uses a topic-specific sentiment summarization for republican and democrat speeches separately. Initial results of experiments conducted using a widely used dataset of US Congress debates~\cite{convote} are encouraging and show that this simple model compares well with classification models that employ state-of-the-art distributional text representations (Section~\ref{sec:exp}). 

\section{Related Work}
\label{sec:relatedWork}

\subsection{Political Ideology Detection}
Political ideology detection has been a relatively new field of research within the NLP community. Most of the previous efforts have focused on capturing the variations in language use in text representing content of different ideologies. Beissmann et al.~\shortcite{ideologyPrediction-text} employ bag-of-word features for ideology detection in different domains such as speeches in German parliament, party manifestos, and facebook posts. Sim et al.~\shortcite{ideological-proportion-speeches} use a labeled corpus of political writings to infer lexicons of cues strongly associated with different ideologies. These ``ideology lexicons'' are then used to analyze political speeches and identify their ideological leanings. Iyyer at al.~\shortcite{rnn-ideology} recently adopted a recursive neural network architecture to detect ideological bias of single sentences. In addition, topic models have also been used for ideology detection by identifying latent topic distributions across different ideologies~\cite{topic-perspective-ideology,topic-model-ideology}. Gerrish and Blei~\shortcite{legislativeRollCalls} connected text of the legislations to voting patterns of legislators from different parties.

\subsection{Sentiment Analysis for Controversy Detection}
Sentiment analysis has proved to be a useful tool in detecting controversial topics as it can help identify topics that evoke different feelings among people on opposite side of the arguments. Mejova et al.~\shortcite{controversy-news2} analyzed language use in controversial news articles and found that a writer may choose to highlight the negative aspects of the opposing view rather than emphasizing the positive aspects of one’s view. Lourentzou et al.~\shortcite{controversy-news3} utilize the sentiments expressed in social media comments to identify controversial portions of news articles. Given a news article and its associated comments on social media, the paper links comments with each sentence of the article (by using a sentence as a query and retrieving comments using BM25 score). For all the comments associated with a sentence, a sentiment score is then computed, and sentences with large variations in positive and negative comments are identified as controversial sentences. Choi et al.~\shortcite{controversy-news} go one step further and identify controversial topics and their sub-topics in news articles. 

\section{Using Topic Sentiments for Ideology Detection}
\label{sec:approach}

Let $\mathcal{D} = \{ \ldots, d, \ldots \}$ be a corpus of political documents such as speeches or social media postings. Let $\mathcal{L} = \{ \ldots, l, \ldots \}$ be the set of ideology class labels. Typical scenarios would just have two class labels (i.e., $|\mathcal{L}| = 2$), but we will outline our formulation for a general case. For document $d\in\mathcal{D}$, $l_d \in\mathcal{L}$ denotes the class label for that document. Our method relies on the usage of topics, each of which are most commonly represented by a probability distribution over the vocabulary. The set of topics over $\mathcal{D}$, which we will denote using $\mathcal{T}$, may be identified using a topic modeling method such as LDA~\cite{blei2003latent} unless a pre-defined set of handcrafted topics is available.

Given a document $d$ and a topic $t$,  our method relies on identifying the sentiment as expressed by content in $d$ towards the topic $t$. The sentiment could be estimated in the form of a categorical label such as one of positive, negative and neutral~\cite{haney2013sentiment}. Within our modelling, however, we adopt a more fine-grained sentiment labelling, whereby the sentiment for a topic-document pair is a probability distribution over a plurality of ordinal polarity classes ranging from {\it strongly positive} to {\it strongly negative}.  Let $\vect{s_{dt}}$ represent the \textit{topic-sentiment} polarity vector of $d$ towards $t$ such that $s_{dt}(x)$ represents the probability of the polarity class $x$. Combining the topic-sentiment vectors for all topics yields a document-specific \textit{topic-sentiment matrix} (TSM) as follows:

\begin{equation}
\vect{S_{d,\mathcal{T}}} = 
\begin{bmatrix}
\ldots & s_{dt_1}(x) & \ldots \\
\ldots & s_{dt_2}(x) & \ldots \\
\vdots & \vdots & \vdots \\
\end{bmatrix}
\label{eq:tsm}
\end{equation}

Each row in the matrix corresponds to a topic within $\mathcal{T}$, with each element quantifying the probability associated with the sentiment polarity class $x$ for the topic $t$ within document $d$. The topic-sentiment matrix above may be regarded as a sentiment signature for the document over the topic set $\mathcal{T}$. 

\subsection{Determining Topic-specific Sentiments}

In constructing TSMs, we make use of topic-specific sentiment estimations as outlined above. Typical sentiment analysis methods (e.g., NLTK Sentiment Analysis\footnote{\url{http://text-processing.com/demo/sentiment/}}) are designed to determine the overall sentiment for a text segment. Using such sentiment analysis methods in order to determine topic-specific sentiments is not necessarily straightforward. We adopt a simple keyword based approach for the task. For every document-topic pair $(t,d)$, we extract the sentences from $d$ that contain at least one of the top-$k$ keywords associated with the topic $t$. We then collate the sentences in the order in which they appear in $d$ and form a mini-document $d_t$. This document $d_t$ is then passed on to a conventional sentiment analyzer that would then estimate the sentiment polarity as a probability distribution over sentiment polarity classes, which then forms the $s_{dt}(.)$ vector. We use $k=5$ and the RNN based sentiment analyzer~\cite{sentiment-rnn} in our method. 

\subsection{Nearest TSM Classification}

We now outline a simple classification model that uses summaries of TSMs. Given a labeled training set of documents, we would like to find the prototypical TSM corresponding to each label. This can be done by identifying the matrix that minimizes the cumulative deviation from those corresponding to the documents with the label. 

\begin{equation}
\vect{S_{l,\mathcal{T}}} = \mathop{\arg\min}_{\vect{X}} \sum_{d \in \mathcal{D} \wedge l_d = l} ||X - \vect{S_{d,\mathcal{T}}}||^2_F
\end{equation}

where $||M||_F$ denotes the Frobenius norm. It turns out that such a label-specific signature matrix is simply the mean of the topic-sentiment matrices corresponding to documents that bear the respective label, which may be computed using the below equation. 

\begin{equation}
\vect{S_{l,\mathcal{T}}} = \frac{1}{|\{d|d \in \mathcal{D} \wedge l_d = l\}|} \sum_{d \in \mathcal{D} \wedge l_d = l} \vect{S_{d,\mathcal{T}}}
\end{equation}

For an unseen (test) document $d'$, we first compute the TSM $\vect{S_{d',\mathcal{T}}}$, and assign it the label corresponding to the label whose TSM is most proximal to $\vect{S_{d',\mathcal{T}}}$. 

\begin{equation}
l_{d'} = \mathop{\arg\min}_l ||\vect{S_{d',\mathcal{T}}} - \vect{S_{l,\mathcal{T}}}||^2_F
\end{equation}

\subsection{Logistic Regression Classification}

In two class scenarios with label such as $\{left, right\}$ or $\{democrat, republican\}$ as we have in our dataset, TSMs can be flattened into a vector and fed into a logistic regression classifier that learns weights - i.e., co-efficients for each topic + sentiment polarity class combination. These weights can then be used to estimate the label by applying it to the new document's TSM. 

\section{Experiments}
\label{sec:exp}

\subsection{Dataset}
We used the publicly available Convote dataset\footnote{\url{http://www.cs.cornell.edu/home/llee/data/convote.html}}~\cite{convote} for our experiments. The dataset provides transcripts of debates in the House of Representatives of the U.S Congress for the year 2005. Each file in the dataset corresponds to a single, uninterrupted utterance by a speaker in a given debate. We combine all the utterances of a speaker in a given debate in a single file to capture different opinions/view points of the speaker about the debate topic. We call this document the view point document (VPD) representing the speaker's opinion about different aspects of the issue being debated. The dataset also provides political affiliations of all the speakers -- Republican (R), Democrat (D), and Independent (I). With there being only six documents for the independent class (four in training, two in test), we excluded them from our evaluation. Table~\ref{tab:data_description} summarizes the statistics about the dataset and distribution of different classes. We obtained $50$ topics using LDA from Mallet\footnote{\url{http://mallet.cs.umass.edu/}} run over the training dataset. The topic-sentiment matrix was obtained using the Stanford CoreNLP sentiment API\footnote{\url{https://nlp.stanford.edu/sentiment/code.html}}~\cite{stanford-corenlp} which provides probability distributions over a set of five sentiment polarity classes.

\begin{table}[]
	\centering
	\begin{tabular}{@{}lrr@{}}
		\toprule
		& \textbf{Training Set} & \textbf{Test Set} \\ 
		\midrule
		\textbf{Republican (R)} & 530 & 194 \\
		\textbf{Democrat (D)} & 641 & 215 \\
		\midrule
		\textbf{Total} & 1175 & 411 \\ \bottomrule
	\end{tabular}
	\caption{Distribution of different classes in the ConVote dataset.}
	\label{tab:data_description}
\end{table}

\subsection{Methods}

In order to evaluate our proposed TSM-based methods - viz., nearest class (NC) and logistic regression (LR) - we use the following methods in our empirical evaluation. 

\begin{enumerate}
	
	\item \textbf{GloVe-d2v:} We use pre-trained GloVe~\cite{glove} word embeddings to compute vector representation of each VPD by averaging the GloVe vectors for all words in the document. A logistic regression classifier is then trained on the vector representations thus obtained.
	
	\item \textbf{GloVe-d2v+TSM:} A logistic regression classifier trained on the GloVe features as well as TSM features. 
\end{enumerate}

\subsection{Results}
\begin{table}[]
	\centering
	\resizebox{\columnwidth}{!}{%
		\begin{tabular}{lccc}
			\toprule
			\textbf{Method} & \textbf{R} & \textbf{D} & \textbf{Total} \\
			\midrule
			\textbf{GloVe d2v} & 0.6391 & 0.6465 & 0.6430 \\
			\textbf{TSM-NC} & {\bf 0.6907} & 0.4558 & 0.5672 \\
			\textbf{TSM-LR} & 0.5258 & {\bf 0.7628} & {\bf 0.6504} \\
			\textbf{GloVe-d2v + TSM} & 0.5051 & 0.7023 & 0.6088 \\
			\bottomrule
		\end{tabular}%
	}
	\caption{Results achieved by different methods on the ideology classification task.}
	\label{tab:classification_results}
\end{table}

Table~\ref{tab:classification_results} reports the classification results for different methods described above. TSM-NC, the method that uses the \tsm vectors and performs simple nearest class classification achieves an overall accuracy of $57\%$. Next, training a logistic regression classifier trained on \tsm vectors as features, TSM-LR, achieves significant improvement with an overall accuracy of $65.04\%$. The word embedding based baseline, the GloVe-d2v method, achieves slightly lower performance with an overall accuracy of $64.30\%$. However, we do note that the per-class performance of GloVe-d2v method is more balanced with about $64\%$ accuracy for both classes. The TSM-LR method on the other hand achieves about $76\%$ for $R$ class and only $52\%$ for the $D$ class. The results obtained are promising and lend weight to out hypothesis that ideological leanings of a person can be identified by using the fine-grained sentiment analysis of the viewpoint a person has towards different underlying topics.

\begin{table*}[]
	\centering
	\begin{tabulary}{0.95\textwidth}{@{}lL@{}}
		\toprule
		\multirow{5}{*}{Most polarizing topics} &  \textbf{H1:} republican congress majority administration leadership n't vote party republicans special   \\
		& \textbf{H2:} administration process vote work included find n't true fix carriers  \\
		& \textbf{H3:} health programs education funding million program cuts care billion year \\
		& \textbf{H4:} health insurance small care coverage businesses plans ahps employees state   \\
		& \textbf{H5:} military center n't students recruiters policy houston men universities colleges \\ 
		\midrule
		\multirow{5}{*}{Least polarizing topics} &  \textbf{L1:} enter director march years response found letter criminal paid general \\
		& \textbf{L2:} corps nuclear year energy projects committee project million funding funds \\
		& \textbf{L3:} osha safety workers commission health h.r employers occupational bills workplace \\
		& \textbf{L4:} gun police industry lawsuits firearms dept chief manufacturers dealers guns \\
		& \textbf{L5:} medal gold medals individuals reagan history legislation ronald king limiting   \\
		\bottomrule
	\end{tabulary}
	\caption{List of most polarizing (top) and least polarizing (bottom) topics as computed using equation~\ref{eq:dist}. }
	\label{tab:topics}
\end{table*}

\subsection{Discussion}

Towards analyzing the significance of the results, we would like to start with drawing attention to the format of the data used in the TSM methods. The document-specific TSM matrices do {\it not} contain any information about the topics themselves, but only about the sentiment in the document towards each topic; one may recollect that $s_{dt}(.)$ is a quantification of the strength of the sentiment in $d$ towards topic $t$. Thus, in contrast to distributional embeddings such as doc2vec, TSMs contain only the information that directly relates to sentiment towards specific topics that are learnt from across the corpus. The results indicate that TSM methods are able to achieve comparable performance to doc2vec-based methods despite usage of only a small slice of informatiom. This points to the importance of sentiment information in determining the political leanings from text. We believe that leveraging TSMs along with distributional embeddings in a manner that can combine the best of both views would improve the state-of-the-art of political ideology detection. 

Next, we also studied if there are topics that are more polarizing than others and how different topics impact classification performance. We identified polarizing topics, i.e, topics that invoke opposite sentiments across two classes (ideologies) by using the following equation.  

\begin{equation}
dist(t,R,D) = ||\vect{s}_{R,t} - \vect{s}_{R,t}||_F
\label{eq:dist}
\end{equation}

Here, $\vect{s}_{R,t} \textnormal{ and } \vect{s}_{D,t} $ represent the sentiment vectors for topic $t$ for republican and democrat classes. Note that these sentiment vectors are the rows corresponding to topic $t$ in TSMs for the two classes, respectively.

Table~\ref{tab:topics} lists the top five topics with most distance, i.e., most polarizing topics (top) and five topics with least distance, i.e.,least polarizing topics (bottom) as computed by equation~\ref{eq:dist}. Note that the topics are represented using the top keywords that they contain according to the probability distribution of the topic. We observe that the most polarizing topics include topics related to healthcare (H3, H4), military programs (H5), and topics related to administration processes (H1 and H2). The least polarizing topics include topics related to worker safety (L3) and energy projects (L2). One counter-intuitive observation is topic related to gun control (L4) that is amongst the least polarizing topics. This anomaly could be attributed to only a few speeches related to this issue in the training set (only 23 out of 1175 speeches mention \textit{gun}) that prevents a reliable estimate of the probability distributions. We observed similar low occurrences of other lower distance topics too indicating the potential for improvements in computation of topic-specific sentiment representations with more data. In fact, performing the nearest neighbor classification $(TSM-NC)$ with only top-10 most polarizing topics led to improvements in classification accuracy from $57\%$ to $61\%$ suggesting that with more data, better \tsm representations could be learned that are better at discriminating between different ideologies.

\section{Conclusions}
We proposed to exploit topic-specific sentiment analysis for the task of automatic ideology detection from text. We described a simple framework for representing political ideologies and documents as a matrix capturing sentiment distributions over topics and used this representation for classifying documents based on their topic-sentiment signatures. Empirical evaluation over a widely used dataset of US Congressional speeches showed that the proposed approach performs on a par with classifiers using distributional text representations. In addition, the proposed approach offers simplicity and easy interpretability of results making it a promising technique for ideology detection. Our immediate future work will focus on further solidifying our observations by using a larger dataset to learn better TSMs for different ideologies. Further, the framework easily lends itself to be used for detecting ideological leanings of authors, social media users, news websites, magazines, etc. by computing their TSMs and comparing against the TSMs of different ideologies.  
\section*{Acknowledgments}
We would like to thank the anonymous reviewers for their valuable
comments and suggestions that helped us improve the quality of this work.

\bibliographystyle{acl_natbib_nourl}
\bibliography{references}

\end{document}